\documentclass[10pt,twocolumn,letterpaper]{article}

\usepackage{cvpr}

\usepackage{graphicx}
\usepackage{amsmath}
\usepackage{amssymb}
\usepackage{booktabs}
\usepackage{color}
\usepackage{multirow}
\usepackage[dvipsnames]{xcolor}
\usepackage{colortbl}
\usepackage{arydshln}
\usepackage{tabularx} 
\usepackage[pagebackref,breaklinks,colorlinks]{hyperref}

\usepackage[capitalize]{cleveref}
\crefname{section}{Sec.}{Secs.}
\Crefname{section}{Section}{Sections}
\Crefname{table}{Table}{Tables}
\crefname{table}{Tab.}{Tabs.}

\begin{document}

\title{Building an Invisible Shield for Your Portrait against Deepfakes}

\author{Jiazhi Guan$^{1,3}$ \quad Tianshu Hu$^{3}$ \quad Hang Zhou$^{3}$ \quad Zhizhi Guo$^{4}$ \\ Lirui Deng$^{1}$ \quad Chengbin Quan$^{1}$ \quad Errui Ding$^{3}$ \quad Youjian Zhao$^{1,2}$\thanks{Corresponding author.}  \\
$^1$Tsinghua University \quad $^2$Zhongguancun Laboratory \quad $^3$VIS, Baidu Inc. \quad $^4$China Telecom \\
{\tt\small \{guanjz20@mails., zhaoyoujian@\}tsinghua.edu.cn}
}

\maketitle

\begin{abstract}
The issue of detecting deepfakes has garnered significant attention in the research community, with the goal of identifying facial manipulations for abuse prevention. Although recent studies have focused on developing generalized models that can detect various types of deepfakes, their performance is not always be reliable and stable, which poses limitations in real-world applications.
Instead of learning a forgery detector, in this paper, we propose a novel framework - Integrity Encryptor, aiming to protect portraits in a proactive strategy.
Our methodology involves covertly encoding messages that are closely associated with key facial attributes into authentic images prior to their public release. Unlike authentic images, where the hidden messages can be extracted with precision, manipulating the facial attributes through deepfake techniques can disrupt the decoding process. Consequently, the modified facial attributes serve as a mean of detecting manipulated images through a comparison of the decoded messages.
Our encryption approach is characterized by its brevity and efficiency, and the resulting method exhibits a good robustness against typical image processing traces, such as image degradation and noise.
When compared to baselines that struggle to detect deepfakes in a black-box setting, our method utilizing conditional encryption showcases superior performance when presented with a range of different types of forgeries. In experiments conducted on our protected data, our approach outperforms existing state-of-the-art methods by a significant margin.
\end{abstract}

\section{Introduction}
\label{sec:intro}

With the rapid development of image synthesis and manipulation \cite{karras2019style,karras2020analyzing,Zhou2021Pose,koujan2020head2head,liu2022semantic}, the concerns of false information and fake news harm the trustworthiness of digital content.
In recent years, there has been a surge of research in the field of deepfake detection \cite{li2018exposing,zheng2021exploring,shiohara2022detecting,wang2021faketagger}. Many of these works attempt to address the problem by learning a hyperplane that distinguishes between binary classes of ``real'' and ``fake'' in high-level feature spaces. While previous methods have demonstrated promising results, their limited generalization capability renders them challenging to apply in real-world scenarios. 
Additionally, training a classifier that relies solely on pre-existing forgery cues may not be sufficient for detecting previously unseen manipulations, posing an ill-posed problem.

\begin{figure}[!t]
\centering
\includegraphics[width=\linewidth]{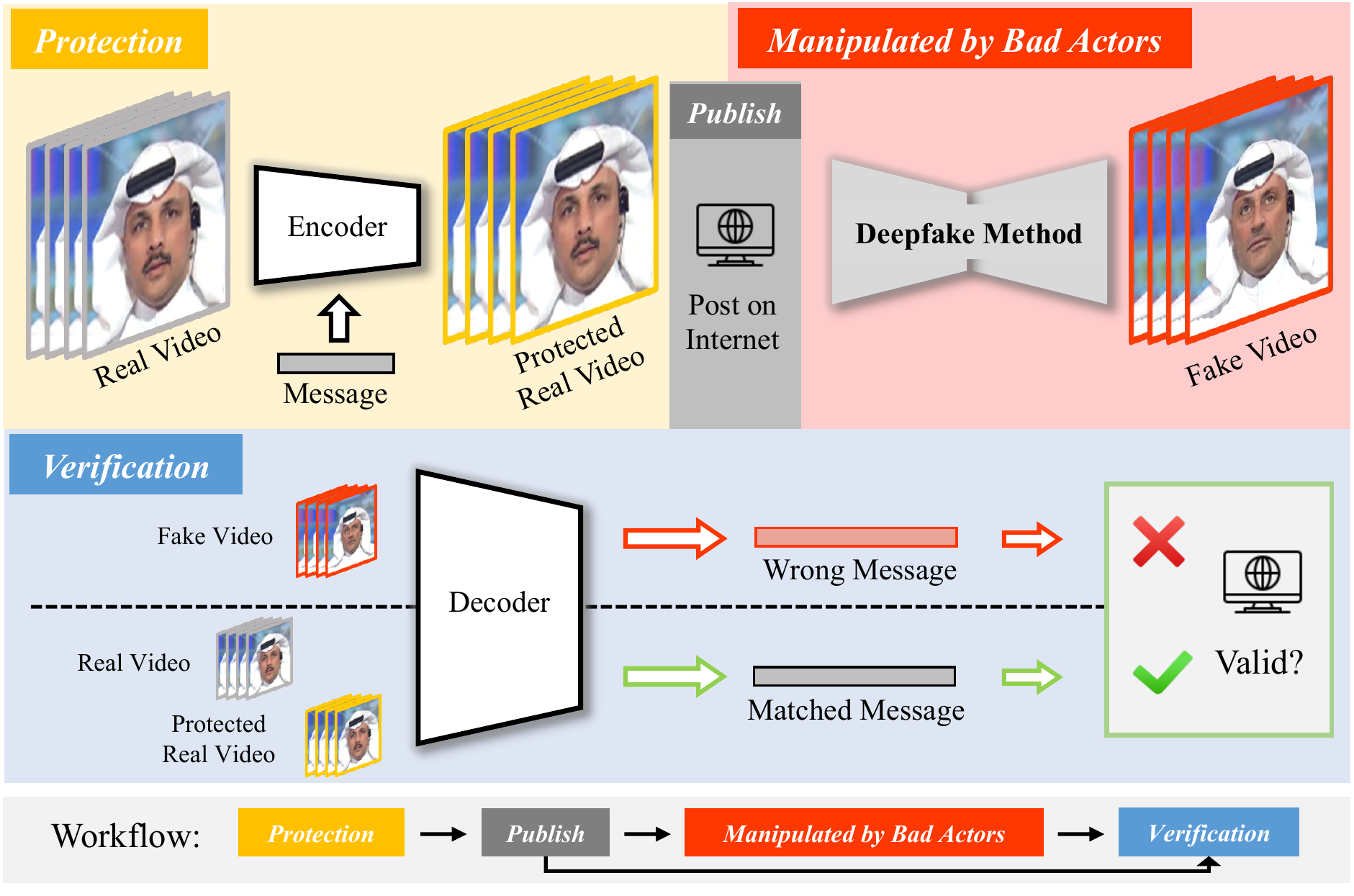}
\caption{
Workflow of the proposed Integrity Encryptor. 
1) {\textbf{\textit{Protection}}}: we embed a secret message (e.g., the hash code of the user name) into the target video that we want to protect. 
2) {\textbf{\textit{Publish}}}: we publish the protected video. 
3) {\textbf{\textit{Manipulation}}}: the video may be manipulated by bad actors. 
4) {\textbf{\textit{Verification}}}: 
if the video undergoes \textit{deepfake-like} modification, the extracted message will not meet the pre-defined criteria, indicating that it is \textit{fake}. On the other hand, if the video is unmodified, we can extract the matched message, indicating that it is \textit{real}.
}
\label{fig:1}
\end{figure}

Alternatively, a distinct area of research has concentrated on preventing abuse towards a particular group, which appears to be a promising direction. We refer to these studies collectively as ``proactive defense'' and classify them into three distinct types.
1) ID-specific validation \cite{agarwal2019protecting,cozzolino2020id}, which is only effective for a limited set of IDs and poses a challenge for extension to a larger ID corpora.
2) Adversarial attack \cite{huang2022cmua,wang2022deepfake,wang2022anti}, which is constrained by the limited transferability of current adversarial attack methods \cite{demontis2019adversarial,xie2021improving}, \textit{i.e.}, they are only reliable confronting accessible deepfake generators, which are used as surrogate models during the training process.
3) Watermark protection \cite{yu2021artificial,yu2020responsible,wang2021faketagger}. It leverages digital watermarking techniques \cite{zhu2018hidden,luo2021dvmark} to safeguard specific samples by embedding secret messages in advance.
However, existing research falls short in ensuring deepfake-agnostic protection. For instance, Wang \textit{et al.} \cite{wang2021faketagger} suggest training an encoder with a deepfake generator acting as a surrogate ``\textit{noiser}'' module, which leads to robustness against ``\textit{noiser}-type'' modifications. However, such practice would yield inconsistent outcomes when faced with manipulations outside of the training corpus.
The same issue arises with \cite{yu2021artificial}, where their method demonstrates efficacy against certain types of deepfake generators, but does not generalize well against others.

To address the need for deepfake-agnostic protection, we introduce a different paradigm (as shown in Fig.~\ref{fig:1}) for proactive defense against deepfakes. 
Our approach differs from \cite{wang2021faketagger} in that it does not assume the inserted messages will remain unchanged after any type of modification, as this assumption is usually impractical.
Instead, our method guarantees that the message is significantly degraded following \textit{deepfake-like} manipulations, while it can still be accurately extracted from the original image even in the presence of perturbations.
By following the verification process outlined in Fig.~\ref{fig:1}, we can readily detect deepfakes. Specifically, if the decoding does not match, it indicates that the sample has undergone \textit{deepfake-like} manipulations.

In order to implement the proposed protection method, the encrypted messages must possess two key properties: being both \textit{noise-robust} and \textit{deepfake-fragile}.
By \textit{noise-robust}, we mean that the message should be accurately extracted even when the image undergoes common perturbations, such as image compression.
By \textit{deepfake-fragile}, we mean that the message should be corrupted or destroyed when the image is subjected to deepfake-like manipulations.
Such idea is akin to semi-fragile watermarking \cite{fridrich1998image,lin2000detection,bolourian2020effective,rey2000blind},
however, the limited robustness to perturbations and the lack of high-level understanding of deepfakes make these methods unable to handle our task.
In addition, recent deep-learning based watermarking studies \cite{zhu2018hidden,luo2021dvmark,luo2020distortion} could guarantee the property of \textit{noise-robust}, but cannot ensure the fragility of the messages against deepfake-like manipulations.
Based on the observation that deepfakes usually change at least one facial attribute such as identity, facial appearance, or mouth movement. We propose a novel framework - Integrity Encryptor to ensure both properties of \textit{noise-robust} and \textit{deepfake-fragile} of the encrypted message.

To ensure the \textit{noise-robust} property, we draw inspiration from previous studies \cite{zhu2018hidden,luo2020distortion} and incorporate a collection of differentiable distortions with a trainable adversarial noise module.
For \textit{deepfake-fragile}, we empower it by injecting messages conditioned on the original facial attributes. So that the correct messages could only be extracted with corresponding facial attributes as a conditional token. By using the input portrait as an invisible shield to hide the message, we can detect if the portrait has been manipulated by bad actors. Any \textit{deepfake-like} changes to the facial attributes will result in a broken conditional token, leading to an incorrectly extracted message and revealing the actions of forgery.

We conduct extensive experiments to evaluate the effectiveness of our proposed method. Firstly, we ensure that the encrypted images maintain high fidelity, which makes it challenging for potential adversaries to suspect tampering. Secondly, we evaluate the robustness of the method against various perturbations to ensure practicality. Most importantly, we evaluate the method's ability to identify deepfakes by testing it on five different types of deepfakes. The results show that our method achieves a detection performance of 97.81\% AUC, indicating its reliability in identifying deepfakes. 
Our contributions can be summarized as:

1) We present a novel solution to the problem of detecting deepfakes, coined Integrity Encryptor, which involves encrypting secret messages that are closely coupled to important facial attributes into portraits. We can thus detect deepfakes by identifying when the extracted message is broken, indicating that the portrait has been tampered with.
2) To achieve \textit{deepfake-fragile} property, we propose to use a conditional token extracted from facial attributes to hide the message. A contrastive loss is further proposed to enhance the invariance of the feature representations in real samples and push away potential fake samples. Therefore, our approach is deepfake-agnostic as it does not require any deepfake simulator during training.
3) With comprehensive experiments, we demonstrate that our method performs favorable compared with related baselines and state-of-the-art (SOTA) detectors. 

\section{Related Works}
\label{sec:related}
A myriad of methods are proposed for deepfake detection in recent years. Most of them are devised following a passive defence routine, \textit{i.e.}, telling the given portrait is real or fake without knowing any prior knowledge of the given portrait. In this section, we first review works belong to this category in Sec.~\ref{sec:Passive Defence for Deepfake} and then introduce works belong to the opposite direction, termed proactive defence, in Sec.~\ref{sec:Proactive Defence for Deepfake}. Lastly, we introduce the related works about watermarking (Sec.~\ref{sec:Watermarking}) and its related application (Sec.~\ref{sec:Content Authentication}).

\subsection{Passive Defence against Deepfake}
\label{sec:Passive Defence for Deepfake}
Since the issues of deepfake abuses are raised, there have been many attempts to identify forgeries following a classification manner. Those methods are trained on given dataset with binary labels (real/fake) and are aimed to spot any coming deepfake. Because all the countermeasures are done after the deepfake is already created, we categorize those methods as passive defence methods. 
Earlier detectors \cite{li2018exposing,afchar2018mesonet,rossler2019faceforensics++} are trained to detect forgeries by learning specific patterns in RGB representations. Zhao \textit{et al.} \cite{zhao2021multi} devise a multi-attentional framework to better capture the local discriminative features. Dang \textit{et al.} \cite{dang2020detection} propose to segment the forged region and classify the authenticity simultaneously. 
With consideration of deepfake generation, where constraints are only employed on the RGB domain, many works propose to explore forgery cues by signal decomposition. \cite{qian2020thinking,gu2021exploiting,li2021frequency} use the DCT responses with deep neural network (DNN) to detect invisiable artifacts. Zhu \textit{et al.} \cite{zhu2021face} further study the better compositions by 3D decomposition. Encouraging results are achieved in those works, however, the generalization ability of detectors is found to be vulnerable.
Many researchers begin to notice the vital problem of cross-dataset generalization. Sun \textit{et al.} \cite{sun2021dual} apply a contrastive learning paradigm to better learn generalized feature representation. Wang \textit{et al.} \cite{wang2021representative} attribute the deficient of generalization to early overfitting to obvious visual artifacts and propose a dynamic data argumentation strategy to enhance the fine-grained forgery detection. Kim \textit{et al.} \cite{kim2021cored} and Tariq \textit{et al.} \cite{tariq2021one} focus on transfer learning or continue learning for generalized and extendable detectors.
Another group of works \cite{masi2020two,zhang2021detecting,gu2022delving,sun2021improving,zheng2021exploring,guan2022delving} particularly focus on the temporal inconsistency of frame-independently generated deepfakes. With the aids of temporal-aware model designs, performance of the detectors is further improved.
In addition to a better feature learning, Li \textit{et al.} \cite{li2018exposing} notice the warping process and propose to identify deepfake by learning universal warping patterns. Similar idea is later studied by Li \textit{et al.} in \cite{li2020face}, they propose to detect deepfake by spotting blending boundary of face swapping. Those ``inevitable'' traces of generation are helpful to the generalization ability, while when the warping and blending artifacts are relieved by advanced deepfake techniques or compression, the performance of those methods degrade considerably. Recent works of \cite{shiohara2022detecting} attempt to learn better forgery patterns from hardly recognized fake samples. 
Although promising generalization results are reported, less than satisfactory performance on challenging dataset \cite{dolhansky2020deepfake} still prohibit the further deployment in the real world.
Recent work of Dong \textit{et al.} \cite{dong2022explaining} reveals that detectors actually learn forgery patterns from artifacts-relevant visual concepts with binary supervision. However, deepfakes generated by different techniques certainly retain different artifacts-relevant visual concepts, which hampers the generalization of all learning-based passive defense methods.

\subsection{Proactive Defence against Deepfake}
\label{sec:Proactive Defence for Deepfake}
To get out of the passive situation of the defensive side in the arms race between forgers and detectors, the researchers explored three types of responses:
1) ID-specific validation \cite{agarwal2019protecting,agarwal2020detecting,cozzolino2020id,dong2022protecting}, 
2) adversarial attack \cite{yeh2020disrupting,ruiz2020disrupting,yang2021defending,huang2022cmua,wang2022deepfake}, 
3) watermark protection \cite{yu2021artificial,yu2020responsible,wang2021faketagger}.
\textit{These works protect only a specific subset of samples with certain prior knowledge (\textit{e.g.}, identity information), we thus categorize them as proactive defense methods. }
For the first type, methods usually learn the person-specific characters and then verify the authenticity by post-comparison on offline validation data. For example, Agarwal \textit{et al.} \cite{agarwal2019protecting} protect the world leaders by learning the personal talking consistency. Cozzolino \textit{et al.} \cite{cozzolino2020id} propose to answer if it is the person who him/her is claimed to be by learning person-specific characters.
For the second type, researchers adopt adversarial attack techniques \cite{goodfellow2014explaining,madry2017towards,poursaeed2018generative} to attack deepfake generator by adding noise in advance. However, most of them ignore the black-box setting. In addition, transferability in adversarial attack \cite{demontis2019adversarial,xie2021improving} remains an open problem, we still cannot ensure the effectiveness of the added noise against untapped generators. 
For the last type, a few works adopt watermarking methods to protect certain samples in advance. Yu \textit{et al.} \cite{yu2020responsible} propose to inject fingerprints into generated deepfakes, thus allowing the bad actor to be tracked should a misuse is happened by user-specific fingerprint matching. However, those fingerprints are added with specific generator during generation, we are certainly not able to interfere bad actors if they create the deepfake themselves. With similar idea, Wang \textit{et al.} \cite{wang2021faketagger} propose to inject user-specific tag before the publish of contents. Their method is trained with specific surrogate deepfake generator, thus the tag could be robust to the modifications made by the generator. Nonetheless, we never have a chance to acknowledge the coming deepfake is created by what generator or what configs, thus black-box evaluation of their method remains concerns.
In a nutshell, existing proactive defense methods are either restricted by limited identity or including surrogate generators in training. In contrast, we propose a new manner to counteract deepfakes without above limitations.

\subsection{Digital Watermarking}
\label{sec:Watermarking}
Digital watermarking is broadly studied \cite{asikuzzaman2017overview,chopra2012lsb,deguillaume1999robust,zhang2019robust}, however, limited robustness and generalization hinders its development. Recently deep learning based methods \cite{zhu2018hidden,luo2021dvmark,lu2021large,tancik2020stegastamp,luo2020distortion,weng2019high,zhang2020udh} are proposed with outstanding performance. HiDDeN \cite{zhu2018hidden} is one of the first attempts to design a DNN for digital watermarking, it shows promising robustness with a noise training strategy. The similar pipeline is later extended in \cite{luo2021dvmark,luo2020distortion} with multi-scale consideration and adversarial training. Works of Lu \textit{et al.} \cite{lu2021large} and Weng \textit{et al.} \cite{weng2019high} focus on the capacity of injected messages. The works of Zhang \textit{et al.} \cite{zhang2019robust}, propose a attention-based framework. Inspired by the attention module, our method is specially designed to address the demands of proactive defense against deepfakes.

\subsection{Content Authentication}
\label{sec:Content Authentication}
Several studies \cite{wolfgang1999fragile,lin1999review,zhang2007statistical} have explored the use of watermarking for forensics purposes. One approach, known as ``fragile watermarking'', involves adding a watermark that becomes undetectable if the protected image is modified in any way \cite{cox2007digital}. However, such methods are not suitable for our task as they can lead to false detections when applied to lossy compressed streams.
A closely related concept, known as ``semi-fragile watermarking'' has been extensively studied for detecting conventional image alterations. Researchers in the field have introduced watermarking techniques that are robust against specific types of lossy compression, such as JPEG compression, by injecting information into hand-crafted domains, such as middle or low frequency AC coefficients in the DCT domain \cite{fridrich1998image,kundur1998towards,lin2000detection,bolourian2020effective}. However,  limited tolerance to specific noise for which they are designed renders them impractical for our task. In contrast, considering both an extensible set of distortions and an adversarial noise module, our method can better withstand various perturbations. 
In contrast to learning appropriate fragile or robust watermarking methods, Rey \textit{et al.} \cite{rey2000blind} propose to concentrate on distinguishing between malicious and non-malicious attacks. However, this approach is limited in its ability to handle deepfakes due to its reliance on only low-level features such as edges and colors. 
On the contrary, the malicious identification ability of our method is empowered by a specifically designed contrastive loss with three facial-related encoders.

\begin{figure*}[!t]
\centering
\includegraphics[width=\linewidth]{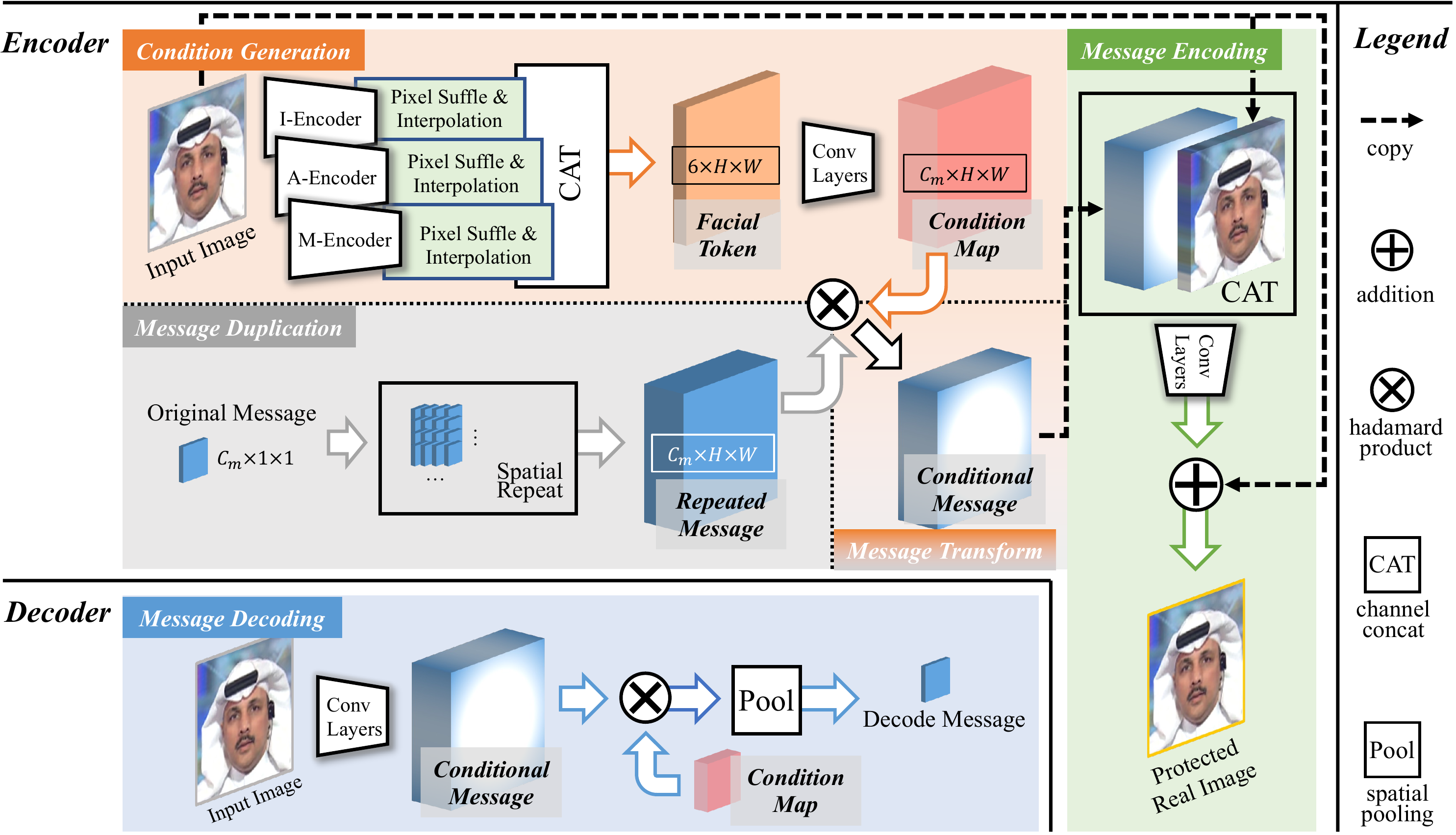}
\caption{
Overview of the proposed Integrity Encryptor.
1) {\textbf{\textit{Condition Generation}}}: using pretrained encoders (I/A/M -Encoder), we extract \textit{Facial Token} of the input, which contains the original facial attributes, and then generate the \textit{Condition Map} based on it.
2) {\textbf{\textit{Message Duplication}}}: we spatially repeat the message.
3) {\textbf{\textit{Message}}} {\textbf{\textit{Transform}}}: we create the \textit{Conditional Message} by multiplication of the \textit{Condition Map} and \textit{Repeated Message}.
5) {\textbf{\textit{Message Encoding}}}: we inject the information into the input. 
6) {\textbf{\textit{Message Decoding}}}: we can only accurately decode the message with the corresponding \textit{Conditional Map}.
} 
\label{fig:2}
\end{figure*}

\section{Method}
\label{sec:method}
In this section, we first introduce the proposed Integrity Encryptor in Sec.~\ref{sec:Integrity Encryptor}, then describe the model optimization in Sec.~\ref{sec:Model Optimization}. Lastly, we explain how to leverage the propose method for deepfake detection in Sec.~\ref{sec:Metrics for Deepfake Detection}.

\subsection{Integrity Encryptor}
\label{sec:Integrity Encryptor}
We illustrate the overview of the proposed Integrity Encryptor in Fig.~\ref{fig:2}. Different parts are denoted with different colors.

\noindent\textbf{Encoder}, which aims to inject a secret message into a given portrait.
In order to entangle the message with the original facial attributes, we first extract the original facial attributes from the input portrait and create the \textit{Facial Token} in the step named \textit{\textbf{Condition Generation}}.
Our method employs the created \textit{Facial Token} as a key to access our message.
In detail, we consider three kinds of crucial facial attributes: identity, facial appearance, and mouth movement, where three pretrained encoders \cite{deng2019arcface,richardson2021encoding,martinez2020lipreading} are used to extract the corresponding embeddings, respectively. 
First, the identity features are extracted using a face recognition model \cite{deng2019arcface}.
Second, considering a pretrained StyleGAN \cite{Karras2019stylegan2} can reconstruct the input face given the corresponding embedding, we argue that the embedding surely retains the information of the original facial appearance.
Therefore, we adopt a StyleGAN encoder \cite{richardson2021encoding} to embed the facial appearance in the latent space of StyleGAN. 
Furthermore, we leverage a lipreading encoder \cite{martinez2020lipreading} to embed the mouth movement.
The above three encoders are kept fixed in our framework (I/A/M -Encoder in Fig.~\ref{fig:2}) and only used to extract the portrait embeddings of the input to form the \textit{Facial Token}. These embeddings are represented by 512-dim tensors, \textit{i.e.}, we have three tensors in shape of $512\times 1\times 1$ for each image. To generate the \textit{Condition Map}, we first upscale the tensors to $2\times 16\times 16$ by pixel shuffle and then interpolate them to the size of input, $2\times H\times W$. The \textit{Facial Token} is created by channel concatenation of these tensors. Then, the \textit{Condition Map} is generated after several convolutional layers for latter usage.
Inspired by the Shannon’s capacity theorem that redundancy is necessary in order to achieve robustness \cite{luo2020distortion}, we learn the spatial repeat operation from \cite{zhu2018hidden} in the \textit{\textbf{Message Duplication}} block. Thus the injecting message is represented as a $C_m\times H\times W$ tensor, where $C_m$ denotes the message length.
In the \textit{\textbf{Message Transform
}} block, we generate the \textit{Conditional Message} by coupling the original message with the \textit{Condition Map}, which is extracted from the original facial attributes as described previously.
The last step in encoder is the \textit{\textbf{Message Encoding}}, which injects the \textit{Conditional Message} into the input image using a lightweight network.

\noindent\textbf{Decoder}.
The decoder simply consists of several convolutional layers and is designed to extract the \textit{Conditional Message} given an input image. In addition, we need to extract the \textit{Condition Map}, following the same pipeline in the encoder processing. Only with a matched \textit{Condition Map}, the original messages can be precisely reconstructed from the \textit{Conditional Message}. This property is specially enabled by a contrastive loss, which we will demonstration in the next section.  

\subsection{Model Optimization}
\label{sec:Model Optimization}
\noindent\textbf{Message Encoding and Decoding.}
We first define the variables and main components of our framework. $x\in \mathbb{R}^{3\times H\times W}$ denotes the input image, $m\in \{0,1\}^{C_m}$ indicates the secret message to be injected, $\operatorname{\textbf{Enc}}$ represents the encoder with trainable parameters $\theta$, and $\operatorname{\textbf{Dec}}$ is the decoder with trainable parameters $\phi$. Given the inputs of $x$ and $m$, $\operatorname{\textbf{Enc}}$ injects the message into the image as $x_{s} = \operatorname{\textbf{Enc}}(x,m; \theta)$; $\operatorname{\textbf{Dec}}$ extracts the message from the watermarked image as $m_o = \operatorname{\textbf{Dec}}(x_{s}; \phi)$. The encoder and decoder are optimized by minimizing
\begin{equation}
    \mathcal{L}_{r}(\theta,\phi) = \operatorname{BCE}(m, m_o),
\end{equation}
where $\operatorname{BCE}$ denotes binary cross entropy loss. To increase the fidelity of $x_{s}$, we softly truncate the differences as $x_{s} = x + \alpha \cdot \operatorname{tanh}(x')$, where $x'$ is the output of convolutions in \textbf{\textit{Message Encoding}} and $\alpha$ is a threshold.

\noindent\textbf{Noise-Robust.}
To enhance the robustness of the injected message, we adopt a noise function $f_{\operatorname{noiser}}$ and an adversarial training strategy. $f_{\operatorname{noiser}}$ includes differentiable approximation of JPEG compression \cite{zhu2018hidden} and gaussian blur. We enhance the robustness of the message by minimizing the cross entropy loss after the noise function as $\min_{\theta,\phi} \mathcal{L}_{n}(\theta,\phi)$,
\begin{equation}
\mathcal{L}_{n}(\theta,\phi) = \operatorname{BCE}(m, \operatorname{\textbf{Dec}}(f_{\operatorname{noiser}}(x_{s}); \phi)).   
\end{equation}
In addition, adversarial training is found to benefit robustness \cite{luo2020distortion}. We further introduce a discriminator $\operatorname{\textbf{Dis}}$ with trainable parameters $\delta_D$ and an adversarial model $\operatorname{\textbf{Adv}}$ with trainable parameters $\delta_A$ in the training phase. $\operatorname{\textbf{Dis}}$ is trained to distinguish between the original image and the watermarked image; $\operatorname{\textbf{Adv}}$ is trained to erase the injected information in $x_{s}$, thus leading to a wrongly decoding. This can be learned from a minimax game $\min_{\{\delta_D,\delta_A\}}\max_{\{\theta,\phi\}} \mathcal{L}_a$, where
\begin{align}
\mathcal{L}_{a} = & \operatorname{\textbf{Dis}}(x;\delta_D) -\operatorname{\textbf{Dis}}(x_{s};\delta_D) \notag \\
& -\operatorname{BCE}(m,\operatorname{\textbf{Dec}}(\operatorname{\textbf{Adv}}(x_{s};\delta_A); \phi)).
\label{eq:adv_loss}
\end{align}

\noindent\textbf{Deepfake-Fragile.} To ensure the injected message can only be extracted with the original facial attributes, we adopt a contrastive learning protocol within a batch of inputs. Given $x_k$ from a batch of image
$\{x_i | i\in \operatorname{S}\}$, where $\operatorname{S} = \{1,2,...,N\}$,
$m_o^k = \operatorname{\textbf{Dec}}(\operatorname{\textbf{Enc}}(x_k, m)|c_k)$ is the extracted message after encoding-decoding, where $c_k$ is the \textit{Condition Map} extracted from $x_k$ in \textbf{Condition Generation}. We take $(m_o^k, m_o^{k'})$ as a positive pair, where
\begin{equation}
m_o^{k'} = \operatorname{\textbf{Dec}}(f_{\operatorname{noiser}}(\operatorname{\textbf{Enc}}(x_k, m))~|~c_k).    
\end{equation}
In contrast, if the decoded message is extracted as 
\begin{equation}
m_o^q = \operatorname{\textbf{Dec}}(\operatorname{\textbf{Enc}}(x_k, m)~|~c_q),    
\end{equation}
where $c_q$ is the \textit{Condition Map} extracted from $x_q$ and $k\ne q$; we take $(m_o^k, m_o^q)$ as a negative pair. Inspired by InfoNCE loss \cite{oord2018representation}, we define the deepfake-fragile loss, $\mathcal{L}_{f}$, as:
\begin{gather}
\mathcal{L}_{f} = \frac{1}{N}  {\textstyle \sum_{k=1}^{N}} \mathcal{L}_{f}^{k}, \label{eq:infonce} \\
\mathcal{L}_{f}^{k} =  -\log\frac{e^{\kappa(m_o^k, m_o^{k'})/\xi}}{e^{\kappa(m_o^k, m_o^{k'})/\xi} + \sum_{q\in (\operatorname{S}-\{k\}) }e^{\kappa(m_o^k, m_o^q)/\xi}},
\end{gather}
where $\kappa(u, v)=\frac{u}{\|u\|}\cdot \frac{v}{\|v\|}$ denotes the cosine similarity and $\xi$ is a temperature parameter. Minimizing $\mathcal{L}_{f}$ will maximize the invariance of the feature representations of the injected message in $x_{s}$, also maximize the distance between $m_o^k$ and $m_o^q$. The latter is closely related to the property of \textit{deepfake-fragile}. Since $m_o^k$ is decoded from $x_{s}$ with the original facial attributes $c_{k}$, it is expected to be able to reconstruct the injected message; while $m_o^q$ is decoded with the facial attributes $c_{q}$ of another image $x_q$ (of another person), the mismatched \textit{Condition Map} will lead to a broken reconstruction. This situation resembles that of deepfakes, which at least change one of the facial attributes we extracted to generate the \textit{Condition Map}, ensuring \textit{deepfake-fragile}. It is worthy mentioning that the whole pipeline is deepfake-agnostic, thus ensuring the generalization ability when untapped types of deepfakes are confronted.

\noindent\textbf{Overall loss.}
So far, we can train our framework by minimizing the overall loss:
\begin{equation}
\mathcal{L} = \mathbb{E}_{x,m}\big [\lambda_r \mathcal{L}_r + \lambda_n \mathcal{L}_n - \lambda_a \mathcal{L}_a + \lambda_f \mathcal{L}_f \big ],
\label{eq:all_loss}
\end{equation}
where $\lambda_\ast$ balance the weights of different parts.

\subsection{Deepfake Detection}
\label{sec:Metrics for Deepfake Detection}

We identify deepfakes by verification as illustrated in the Fig.~\ref{fig:1}. Without loss of generality, the injected message is represented by a binary string with fixed length. Thus the correctness of the verification can be evaluated by {Bit Error Rate} (BER). The smaller the BER, the more accurate the decoded information is. In contrast, the larger the BER, the more likely it is a deepfake. Till now, we still have a problem of how to determine the threshold for distinguishing fake from real. Here we provide two protocols for white-box and black-box evaluations, respectively.

For \textbf{white-box evaluation}, we assume that we can access to a validation set where the fake samples are created by the same deepfake generator as the test set. This setting is in line with some real-world situations. For example, bad actors may use open-source generators to make deepfakes, and we can access these models to determine an appropriate threshold in advance. We can determine the threshold $\tau$ as:
\begin{align}
\tau = \max_{\tau} \bigg (& \mathbb{E}_{x_{\mathrm{fake}}\sim S_V} \big [\mathbf{P}(\mathrm{fake}|x_{\mathrm{fake}};\tau) \big ]  \notag \\
&- \mathbb{E}_{x_{\mathrm{real}}\sim S_V}\big [\mathbf{P}(\mathrm{fake}|x_{\mathrm{real}};\tau) \big ] \bigg ),
\label{eq:white-box tau}
\end{align}
where $x_{\mathrm{real}}$ and $x_{\mathrm{fake}}$ are sampled from the validation set $S_V$, and $\mathbf{P}$ is defined as:
\begin{equation}
\mathbf{P}(\mathrm{fake}|x;\tau) = \begin{cases}
~~~~~~~\frac{0.5}{\tau}  & \text{ if } r^{x} \le \tau, \\
0.5 + \frac{0.5*(r^{x}-\tau)}{0.5-\tau}   & \text{ if } \tau <  r^{x} \le 0.5, \\
~~~~~~~~~1  & \text{ if }  0.5 < r^{x},
\end{cases}
\label{eq:fake_p}
\end{equation}
where $r^{x}$ denotes the BER of $x$, and $\mathbf{P}(\mathrm{fake}|x;\tau)$ represents the probability that $x$ is \textit{fake}.
However, in more cases, we cannot get fake samples of the same origin as the bad actors in advance, which we regard as black-box evaluation.

For \textbf{black-box evaluation}, we have only the real data. To apply the deepfake detection, the threshold we set should not lead to misclassification of the real sample, even if the real sample is disturbed to some extent (e.g., compression). Therefore, the threshold should be at least greater than the BER obtained by the robust evaluation (Sec. \ref{sec:Robustness}). 

For both the protocols, once the threshold is determined, we can predict the fake-probability of the input by Eq.~\ref{eq:fake_p}. Then, we can calculate binary classification accuracy (ACC) and Area Under the Receiver Operating Characteristic Curve (AUC) as evaluation metrics to intuitively demonstrate the performance of deepfake detection.

\section{Experiments}
\label{sec:exp}

\subsection{Set Up}
\label{sec:Set Up}

\noindent\textbf{Dataset.}
Following many previous works in deepfake detection, we use the popular dataset FF++ \cite{rossler2019faceforensics++} for evaluations. Since we need to inject the message before modifications, only 1000 real videos are used. For train/val/test split, we follow the original setting in \cite{rossler2019faceforensics++}, \textit{i.e.}, 720 videos for training, 140 videos for validation, and 140 videos for testing.

\noindent\textbf{Deepfake Models.}
For deepfake generation, we include the three types most likely to be maliciously exploited: 1) face swap, 2) appearance modification (face edit), and 3) lip/expression driven (face reenact). 
The first type, which is the most well known category of deepfake abuses. We include two models, the first of which, \textbf{Swap}, is simply implemented through nearest neighbour search of landmarks and facial blending; the second is a SOTA generator, \textbf{SimSwap} \cite{chen2020simswap}. 
For the second type of appearance modification, we use \textbf{StarGAN} \cite{choi2018stargan} and \textbf{AttGAN} \cite{he2019attgan}. Both of the two models are able to change some attributes of the face, such as age, hair, etc. 
For the last type termed lip/expression driven, it can be easily overlooked but can be extremely powerful. We have witnessed many national leaders being manipulated to make speeches that they would never have made.
To simulate such abuse, we adopt 
\textbf{Wav2Lip} \cite{prajwal2020lip} to drive the lip by random audio. 
It is important to note that the deepfake models used for evaluation purposes are completely independent from the training of our proposed method.

\noindent\textbf{Compared Methods.}
We include both works of digital watermarking and deepfake detection for comparisons. In the case of the former, we use one traditional method, \textbf{LSB} \cite{chopra2012lsb} and two DNN-based methods, \textbf{HiDDeN} \cite{zhu2018hidden} and \textbf{Riva} \cite{zhang2019robust}. LSB encode messages by replacing the lowest-order bits of each image pixel with the bits of the messages. HiDDeN and Riva are designed following an encoder-decoder way. 
For deepfake detection, we include five detectors. 
\textbf{Xception} \cite{chollet2017xception} is popular backbone used for deepfake detection.
\textbf{SBI} \cite{shiohara2022detecting} is one of the SOTA detectors trained using self-blending data.
We use \textbf{R3D} \cite{hara2017learning} as a temporal-aware baseline.
\textbf{FTCN-TT} \cite{zheng2021exploring} is proposed to learn the temporal coherence by 3D convolutions with spatial receptive field of $1\times1$. 
\textbf{LTTD} \cite{guan2022delving} is a recent work focusing on low-level temporal learning with Transformer blocks. 

\noindent\textbf{Message Encoding/Decoding}.
Without loss of generality, message used in our experiments is represented by a 32-bit string. Thus the decoding accuracy and BER can be calculated by bit-by-bit comparison. The image size for encoding/decoding is $128\times 128$.

\noindent\textbf{Implementation Details}. 
``Conv Layer'' of the proposed {Integrity Encryptor} (Fig.~\ref{fig:2}) is composed of two convolutional layers with a kernel size of $7\times 7$. Discriminator used for adversarial training ($\operatorname{\textbf{Dis}}$ in Eq.~\ref{eq:adv_loss}) is composed of three $3\times 3$ convolutional layers with down sample. And the adversarial model ($\operatorname{\textbf{Adv}}$ in Eq.~\ref{eq:adv_loss}) is composed of two layers of $3\times 3$ convolutional layers.
Temperature parameter $\xi$ in Eq.~\ref{eq:infonce} is set to 0.5. $\lambda_r, \lambda_n, \lambda_a, \lambda_f$ in Eq.~\ref{eq:all_loss} are set to 1, 1, 0.1, and 1, respectively. Our models are trained using Adam optimizer with a learning rate of $10^{-3}$. All the experiments are conducted with four NVIDIA 3090 GPUs.

\begin{table}[b]
\centering
\resizebox{.8\linewidth}{!}{
\begin{tabular}{lccc}
\toprule
Methods & PSNR $\uparrow$ & LPIPS $\times 100 \downarrow$ & SSIM $\uparrow$  \\
\hline
\noalign{\smallskip}
LSB \cite{chopra2012lsb} & {54.1295} & 0.4563 & {0.9962} \\
HiDDeN \cite{zhu2018hidden} & 37.2998 & 0.7300 & 0.9829 \\
Riva \cite{zhang2019robust} & 42.5087 & 0.3317 & {0.9950} \\
Ours & {42.5623} & {0.3114} & 0.9916 \\
\bottomrule
\end{tabular}
}
\caption{Fidelity evaluation.}
\label{tab:Fidelity}
\end{table}

\begin{table}[b]
\centering
\resizebox{.9\linewidth}{!}{
\begin{tabular}{lcccc}
\toprule
                        & LSB \cite{chopra2012lsb} & HiDDeN \cite{zhu2018hidden} & Riva \cite{zhang2019robust} & Ours \\
\hline
\noalign{\smallskip}
\multicolumn{1}{c}{BER $\downarrow$} & 0 & 0.1107 & 0.0275 & 0.0176 \\
\bottomrule
\end{tabular}
}
\caption{Message extraction. The smaller the BER, the less information is lost after the encoding-decoding processing.}
\label{tab:Accuracy}
\end{table}

\begin{table}[b]
\centering
\resizebox{\linewidth}{!}{
\begin{tabular}{cccccc}
\toprule
\begin{tabular}[c]{@{}c@{}}Perturbation$\rightarrow$\\ Level$\downarrow$ \end{tabular}& \begin{tabular}[c]{@{}c@{}}Compression\\ \textit{quality}\end{tabular} & \begin{tabular}[c]{@{}c@{}}Down Scale\\ \textit{ratio}\end{tabular} & \begin{tabular}[c]{@{}c@{}}Gaussian Blur\\ \textit{kernel}\end{tabular} & \begin{tabular}[c]{@{}c@{}}Gaussian Noise\\ \textit{variance}\end{tabular} & \begin{tabular}[c]{@{}c@{}}Random Drop\\ \textit{num. hole}\end{tabular} \\
\hline
\noalign{\smallskip}
0 & \multicolumn{5}{c}{\textit{Clean Data}}    \\
\hdashline
\noalign{\smallskip}
1 &   90   & 0.9     &  3    & 10      &  2    \\
2 &   80   & 0.8     &  5    & 20     &   3   \\
3 &   70   & 0.7     &  7    & 30      &  4   \\
4 &   60   & 0.6     &  9    &  40     &  5    \\
5 &   50   & 0.5     &  11    &  50     &  6    \\
\bottomrule
\end{tabular}
}
\caption{Parameters of five perturbations at six levels.}
\label{tab:noise_params}
\end{table}

\begin{figure}[b]
\centering
\includegraphics[width=\linewidth]{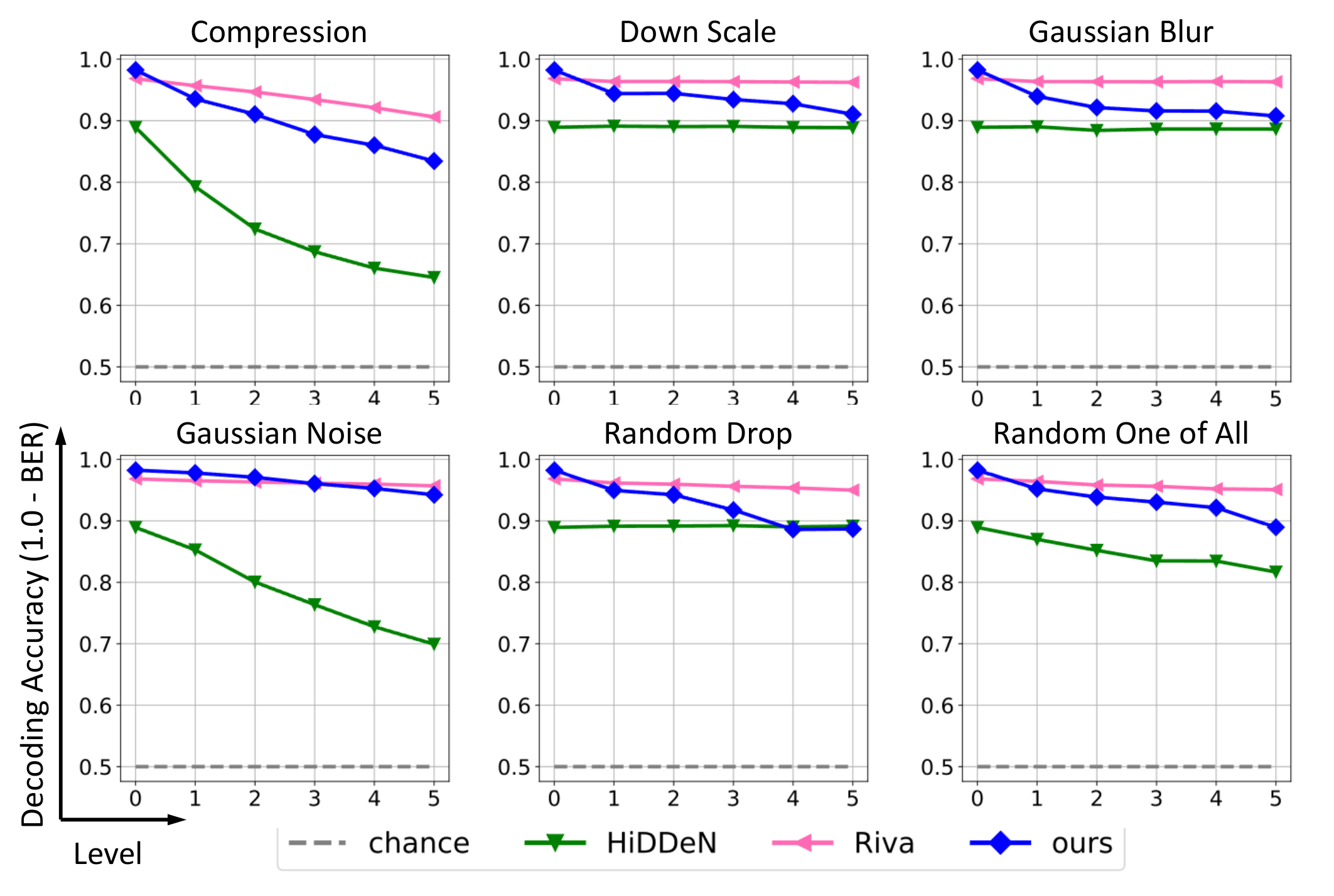}
\caption{Roubustness evaluation. ``Random One of All'' denotes that we randomly choose one of other five perturbations. ``chance'' denotes the accuracy of random guess.}
\label{fig:3}
\end{figure}

\begin{table*}[!th]
\centering
\resizebox{\linewidth}{!}{
\begin{tabular}{lcccccc}
\toprule
\multirow{3}{*}{Methods} & \multicolumn{6}{c}{ACC $\uparrow$ / AUC $\uparrow$ (\%)}               \\ 
\cline{2-7} 
\noalign{\smallskip}
 & \multicolumn{2}{c}{\textit{Face Swap}} & \multicolumn{2}{c}{\textit{Face Edit}} & \textit{Face Reenact} & \multirow{2}{*}{\textit{\textbf{Average}}} \\ 
\cline{2-6}
\noalign{\smallskip}
                         & Swap & SimSwap & AttGAN & StarGAN & Wav2Lip &  \\ 
\hline
\noalign{\smallskip}
Xception \cite{chollet2017xception}& 46.79/47.28 & 58.93/63.07 & 42.50/40.69 & 49.29/52.47 & 68.93/75.96 & 53.29/55.89  \\
R3D-34 \cite{hara2017learning}& 62.50/79.29 & 51.43/63.08 & 46.43/48.26 & 48.57/59.35 & 53.21/77.27 & 52.43/65.45 \\
FTCN-TT \cite{zheng2021exploring} & 96.79/99.87 & 50.36/40.64 & 59.29/73.71 & 55.00/75.28 & 53.57/63.61 & 63.00/70.62 \\
SBI \cite{shiohara2022detecting}& 56.07/54.02 & 63.93/76.70 & 71.79/85.78 & 77.14/89.20 & 78.93/89.54 &  69.57/79.05 \\
LTTD \cite{guan2022delving}& 63.21/89.42 & 51.43/70.96 & 51.79/72.70 & 53.57/73.56 & 65.00/86.47 & 57.00/78.62 \\

\hdashline
\noalign{\smallskip}

HiDDeN \dag \cite{zhu2018hidden} & 51.79/53.34 & 52.50/51.75 & 49.29/51.26 & 53.21/53.76 & 52.14/49.93 & 51.79/52.01 \\
Riva \dag \cite{zhang2019robust} & 81.42/89.95 & 83.21/90.23 & 67.86/76.10 & 86.07/92.12 & 52.14/53.30 & 74.14/80.34 \\
\rowcolor[HTML]{D9D3D3}
Ours \dag & 99.29/100 & 85.36/92.01 & 93.93/97.51 & 94.64/99.59 & 99.29/100 & 94.50/97.82 \\

\noalign{\smallskip}
\hdashline
\noalign{\smallskip}

HiDDeN \ddag \cite{zhu2018hidden} & 50.36/55.41 & 50.00/52.16 & 50.71/51.26 & 51.43/53.76 & 49.29/49.93  & 50.36/52.50 \\ 
Riva \ddag \cite{zhang2019robust} & 82.50/89.92 & 52.50/56.11 & 65.71/71.76 & 81.07/89.99 & 51.07/53.66 & 66.57/72.29 \\ 
\rowcolor[HTML]{D9D3D3}
Ours \ddag & {98.21}/{100} & 77.50/{91.96} & {87.86}/{97.51} & {94.64}/{99.59} & {99.29}/{100} & {91.50}/{97.81} \\ 

\bottomrule
\end{tabular}
}
\caption{Performance of deepfake detection. We compare the proposed proactive defense strategy with SOTA detectors, as well as watermarking baselines. \dag: white-box evaluation, \ddag: black-box evaluation.}
\label{tab:Detection}
\end{table*}

\begin{figure}[!h]
\centering
\includegraphics[width=\linewidth]{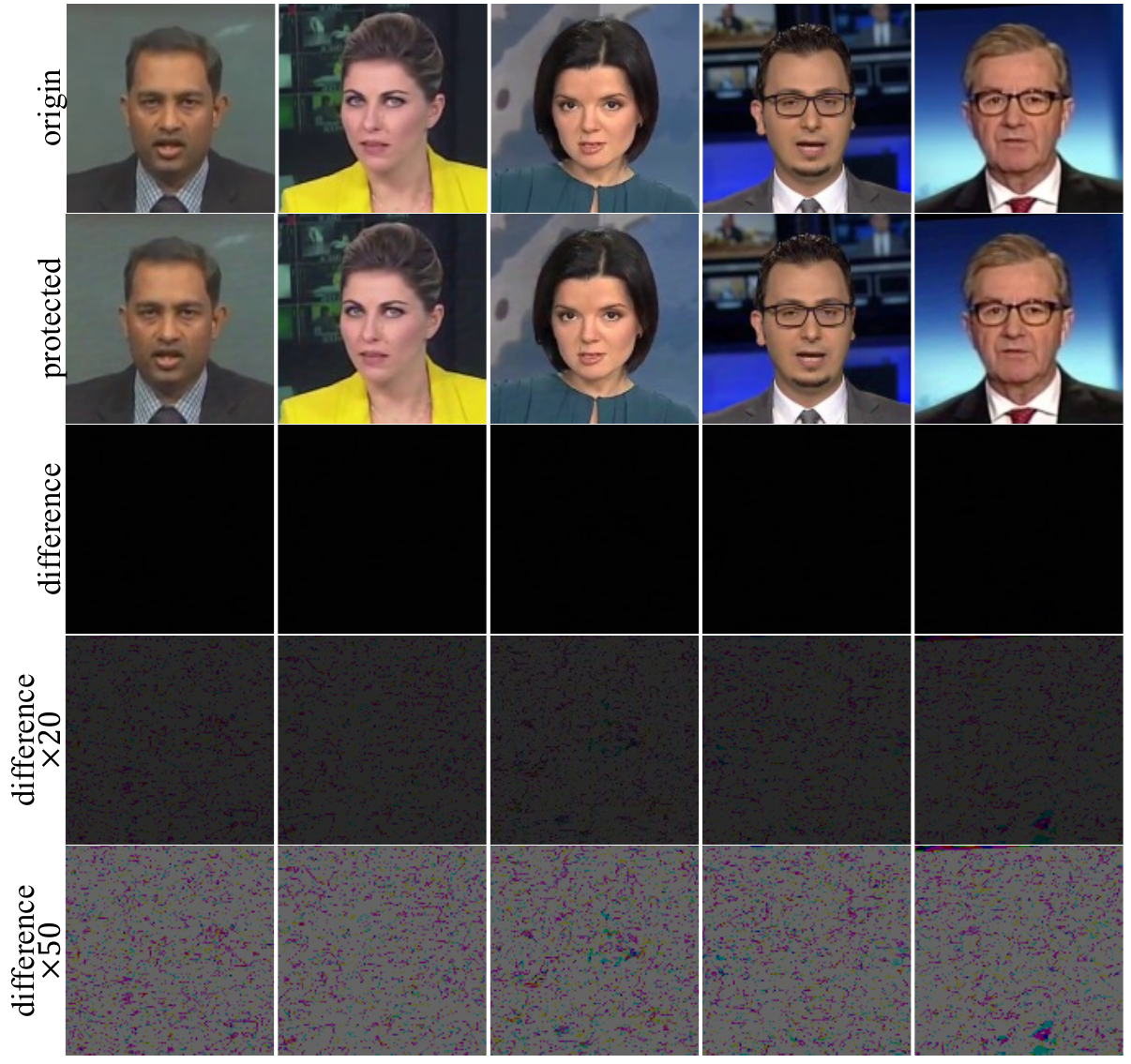}
\caption{Visualization of injected message, which has little side-effect on the visual quality.}
\label{fig:4}
\end{figure}

\subsection{Fidelity}
\label{sec:Fidelity}
The injected message should have negligible side effect on the visual conception of the original image, thus ensuring the original information can be accurately delivered and prohibiting the potential adversary's suspect. In this section, we evaluate the fidelity of the results using several commonly used metrics, including Peak Signal-to-Noise Ratio (PSNR), LPIPS \cite{zhang2018perceptual}, and Structural Similarity Index Measure (SSIM).

We train the models on 720 real videos of the train set, and conduct evaluation on the test set. From Table \ref{tab:Fidelity}, our method shows competitive performance in terms of fidelity. We further intuitively demonstrate the results in Fig.~\ref{fig:4}. 
The message injected has little side-effect on the visual quality; no specific pattern could be identified across different protected samples. Thus the samples we protected are almost indistinguishable from the original ones.

\noindent\textbf{Dose high fidelity leads to low accuracy?} With such little impact on the visual content, it is questionable whether the injected message can be extracted precisely. We tabulate the comparisons in Table \ref{tab:Accuracy}. Our method shows good performance in terms of the information integrity.

\subsection{Robustness}
\label{sec:Robustness}
The injected message should be robust to common post-processing. This is a prerequisite for our approach to work properly, preventing a corrupt message decoded from a real sample from causing false positive (misclassification of real sample as fake). In this section, we evaluate our method with five kinds of perturbations including image compression, down scale, gaussian blur, gaussian noise, and random dropout. Comparisons with HiDDeN and Riva are illustrated in Fig.~\ref{fig:3}. We report the decoding accuracy on the test set under five types of perturbations at six levels. Details of the perturbations are tabulated in Table~\ref{tab:noise_params}. 
From the results, our method achieves SOTA robustness compared with the baselines. As for HiDDeN, image compression and gaussian noise have a significant impact on decoding accuracy. While Riva keeps a high accuracy consistently through all kinds of perturbations. Our method extracts message depending on certain conditional features extracted from the input, when these high-level information is interfered considerably, the decoding accuracy will decrease. Note that we also conduct experiments with LSB. LSB performs perfectly on the clean data, but its performance on perturbed data is no different from random-guess result. \textit{We thus omit LSB in following experiments.}

In a nutshell, our method shows excellent performance with a decoding accuracy of 98.24\% and good robustness to several kinds of perturbations. The BER evaluated by random applying the highest-level perturbations is $0.1105$, which we denote as $\tau_{b}$. $\tau_{b}$ will be the threshold for black-box evaluation in Sec.~\ref{sec:Detection}, since the BER of real sample, albeit distorted, will not exceed $\tau_{b}$ in most cases, greatly avoiding case of false positive.

\subsection{Deepfake Detection}
\label{sec:Detection}
Once the injected message can be adequately extracted, we can accurately identify deepfakes. 
Due to the strong coupling between the injected message and the original facial attributes, any partial or complete destruction of these attributes during the generation of deepfakes results in a mismatch in decoding.
In this section, we conduct experiments against five deepfake models as described in Sec.~\ref{sec:Set Up}. For comparison, we also train five passive detectors on Face2Face of FF++ (except SBI, which is trained using self-created data). The results are illustrated in Table~\ref{tab:Detection}.

\noindent\textbf{Passive detectors.}
For the first part of the five passive detectors, We see that even SOTA detectors struggle to generalize to untapped types of forgery. In addition, we find that different detectors perform very differently on different types of forgery. For example, FTCN-TT spots forgeries of Swap perfectly but cannot handle DNN-based forgeies of SimSwap. In general, passive detectors always tend to fit on specific forgery cues and will be hard to generalize to unseen types of deepfakes. This is also the key factor that hinders the deployment of deepfake detection system. It should be noted that a direct comparison between passive detectors and proactive defenders may be unfair. Nevertheless, we present these results to illustrate a point that passive detectors could be ineffective when it comes to detecting deepfakes of various kinds.

\noindent\textbf{White-box proactive defence.}
For the second part, the three rows show a comparison between two watermarking baselines and our method evaluated on the white-box proactive defence strategy. We first train the models using only the real data, thus being agnostic to deepfakes. Then, we solve Eq.~\ref{eq:white-box tau} to determine the value of $\tau$. Therefore, given a sample $x$, we can calculate the fake probability by Eq.~\ref{eq:fake_p}. To solve Eq.~\ref{eq:white-box tau}, we simply get an approximate solution by grid search. For HiDDeN, its performance is no different from random guess. For robustness consideration, HiDDeN replicates the message all over the whole image. When the face is partial modified, the original information can also be extracted from the duplicated ones, thus leading to a collapse with our strategy. As for Riva, it shows good performance on several types of deepfakes. This is closely related to the attention module of Riva. Since it inject the message only in the attended parts of the image, when these regions are happened to be modified, the broken decoding result will indicate deepfake.
As previously mentioned in Section~\ref{sec:Robustness}, LSB performs no differently than random guess when applied to deepfakes and perturbed data. Therefore, detection results using LSB are meaningless since even the perturbed real images will be classified as fake. Consequently, we omitted LSB in this part of the analysis.
In contrast to the white-box protocol, we typically encounter the scenario of a black-box attack where we cannot access to fake samples in advance. In this situation, we further evaluate our method under the black-box protocol.

\noindent\textbf{Black-box proactive defence.}
For this part, we set the threshold $\tau$ as the BER $\tau_b$ evaluated in Sec.~\ref{sec:Robustness}, thus at least, the real sample will not be misclassified. We show the results in the bottom rows of Table~\ref{tab:Detection}, our method performs basically the same as the white-box scenario on the black-box protocol. This benefits from the conditional injection designs of our model. Broken \textit{Condition Map} will lead to negative validation when the original facial attributes change. In contrast, Riva does not perform better than the passive detectors. The comparison reveals the vital importance of ensuring both \textit{noise-robust} and \textit{deepfake-fragile}. Watermarking methods can guarantee only the former but not the latter. Our method, featuring delicate conditional injection designs, demonstrates promising detection capabilities against deepfakes while being agnostic to the specific types of deepfakes.

\subsection{Ablations}

\noindent\textbf{Noise-Robust.} 
As introduced in Sec.~\ref{sec:Model Optimization}, the robustness of the injected message is enhanced by adopting a noise function $f_{\operatorname{noiser}}$ and an adversarial training strategy ($\operatorname{Adv}$ in Eq.~\ref{eq:adv_loss}). In this section, we conduct experiments on the above two settings to investigate their impact on the property of \textit{noise-robust}. We train our model without either of the two settings and evaluate on perturbed data at level-3 (described in Table.~\ref{tab:noise_params}). The results are tabulated in Table.~\ref{tab:ablation_noise_robust}.
Compared with the results obtained using the model aided by both $f_{\operatorname{noiser}}$ and $\operatorname{Adv}$, removing either one of these measures leads to a degradation in the model's robustness against perturbations.

\noindent\textbf{Deepfake-Fragile.} 
To ensure the important property of \textit{deepfake-fragile}, we introduce the loss described in Eq.~\ref{eq:infonce}. By constructing negative pairs of secret message and \textit{Conditional Map} generated by another person, Eq.~\ref{eq:infonce} is able to induce broken reconstruction when facing a forged portrait. In this section, we validate the effects of Eq.~\ref{eq:infonce} by adjusting the loss weight $\lambda_f$ in Eq.~\ref{eq:all_loss}. Deepfake detection results (black-box evaluation) are tabulated in Table.~\ref{tab:ablation_deepfake_fragile_loss}.
It is evident that the introduced {deepfake-fragile} loss (Eq.~\ref{eq:infonce}) greatly enhances the invariance of the feature representations in real samples, leading to reliable detection of broken decoding in deepfakes.

\noindent\textbf{Conditional Injection.} 
Removing the steps of \textbf{\textit{Condition Generation}} and \textbf{\textit{Message Transform}} results in the injected message no longer being coupled to the original facial attributes, thereby causing the entire protection scheme to fail.
Under the black-box protocol, the performance averaged from five types of deepfakes drops from 91.50\%/97.81\% to 64.79\%/70.45\% (ACC/AUC).

\begin{table}[]
\centering
\resizebox{\linewidth}{!}{
\begin{tabular}{cccccc}
\toprule
\multirow{2}{*}{Models} & \multicolumn{5}{c}{BER $\downarrow$} \\
\cline{2-6}
\noalign{\smallskip}
                  & Compression   & \begin{tabular}[c]{@{}c@{}}Down\\ Scale\end{tabular}   & \begin{tabular}[c]{@{}c@{}}Gaussian\\ Blur\end{tabular}    & \begin{tabular}[c]{@{}c@{}}Gaussian\\ Noise\end{tabular}    & \begin{tabular}[c]{@{}c@{}}Random\\ Drop\end{tabular}   \\
\hline
\noalign{\smallskip}
Ours &  0.1225  & 0.0656  &  0.0842  & 0.0395 & 0.0824 \\
Ours w/o $f_{\operatorname{noiser}}$ & 0.1490 & 0.1012 &  0.1146 &  0.0760 & 0.1162   \\
Ours w/o $\operatorname{Adv}$ & 0.2097 &  0.0913  & 0.0923 & 0.1066  & 0.0876  \\
\bottomrule
\end{tabular}
}
\caption{Ablation on noise-robust.}
\label{tab:ablation_noise_robust}
\end{table}

\begin{table}[]
\centering
\resizebox{\linewidth}{!}{
\begin{tabular}{ccccc}

\toprule
\multirow{2}{*}{$\lambda_f$} & \multicolumn{4}{c}{ACC $\uparrow$ / AUC $\uparrow$ (\%)}               \\ 
\cline{2-5} 
\noalign{\smallskip}
                         & SimSwap & StarGAN & Wav2Lip & \textbf{\textit{Average}}  \\ 
\hline
\noalign{\smallskip}
0.0 & 65.00/67.02 & 87.14/97.04 & 87.50/98.04 & 79.88/87.37 \\
0.1 & 76.43/77.20 & 93.93/99.39 & 89.64/98.30 & 86.67/91.63 \\
1.0 & 77.50/91.96 & 94.64/99.59 & 99.29/100 & 90.48/97.18 \\
10.0 & 71.43/87.93 & 92.50/96.98 & 90.71/96.25 & 84.88/93.72 \\
\bottomrule
\end{tabular}
}
\caption{Ablation on deepfake-fragile.}
\label{tab:ablation_deepfake_fragile_loss}
\end{table}

\section{Conclusion}
\label{sec:conclusion}
In this paper, we present a proactive defense strategy against the malicious use of deepfakes. We propose a novel Integrity Encryptor that ensures the two crucial properties of being \textit{noise-robust} and \textit{deepfake-fragile}. Through experiments conducted on various types of deepfakes, our method demonstrates promising results.
For practical applications, users who wish to protect their portraits on social media platforms, such as Twitter, must do so before the first upload. This feature makes our method suitable for service providers, as they can easily protect the portraits uploaded by users before publicly posting them on their platform.

Due to the limited generalization ability of passive detectors to unseen deepfakes, we suggest exploring different defense strategies, like proactive defender proposed in this study, for addressing the problem of deepfakes abuse. We believe that further research in this direction will lead to more diverse and effective solutions.

{\small
\bibliographystyle{ieee_fullname}
\bibliography{mylib}
}

\end{document}